\documentclass[lettersize,journal]{IEEEtran}

\usepackage{algorithm}
\usepackage{algpseudocode}
\usepackage{graphicx}
\usepackage{multirow}
\usepackage{subfigure}
\usepackage{ifpdf}
\usepackage{diagbox}
\usepackage{bbm}
\usepackage{amssymb}  
\usepackage{pifont}   
\usepackage{bbding}   
\usepackage{booktabs}
\usepackage{makecell} 
\usepackage{colortbl}  
\usepackage{xcolor}
\usepackage{float}
\usepackage{acronym}
\usepackage{amsmath}
\usepackage{url}
\definecolor{pinkkkk}{RGB}{255,102,255}
\newcolumntype{I}{!{\vrule width 1pt}}  

\begin{document}

\title{Edge Approximation Text Detector}

\author{
	Chuang~Yang,
	Xu~Han,
	Tao~Han,
	Han~Han,
	Bingxuan~Zhao,
	and Qi~Wang,~\IEEEmembership{Senior Member,~IEEE
	}
	
	\thanks{
		Chuang~Yang, Xu~Han, Han~Han, Bingxuan~Zhao, and Qi~Wang are with the School of Artificial Intelligence, OPtics and ElectroNics (iOPEN), Northwestern Polytechnical University, Xi'an 710072, P.R. China. 
		
		Tao~Han is with Shanghai Artificial Intelligence Laboratory, Longwen Road 129, Xuhui District, 200232 Shanghai, China.
		
		E-mail: omtcyang@gmail.com, hxu04100@gmail.com, hantao10200@gmail.com, 1356376210@qq.com, bxuanzhao202@gmail.com, crabwq@gmail.com.}

	\thanks{Qi~Wang is the corresponding author.}
	
}

\markboth{}%
{Shell \MakeLowercase{\textit{et al.}}: A Sample Article Using IEEEtran.cls for IEEE Journals}


\maketitle

\begin{abstract}
Pursuing efficient text shape representations helps scene text detection models focus on compact foreground regions and optimize the contour reconstruction steps to simplify the whole detection pipeline. Current approaches either represent irregular shapes via box-to-polygon strategy or decomposing a contour into pieces for fitting gradually, the deficiency of coarse contours or complex pipelines always exists in these models. Considering the above issues, we introduce \textit{EdgeText} to fit text contours compactly while alleviating excessive contour rebuilding processes. Concretely, it is observed that the two long edges of texts can be regarded as smooth curves. It allows us to build contours via continuous and smooth edges that cover text regions tightly instead of fitting piecewise, which helps avoid the two limitations in current models. Inspired by this observation, EdgeText formulates the text representation as the edge approximation problem via parameterized curve fitting functions. In the inference stage, our model starts with locating text centers, and then creating curve functions for approximating text edges relying on the points. Meanwhile, truncation points are determined based on the location features. In the end, extracting curve segments from curve functions by using the pixel coordinate information brought by truncation points to reconstruct text contours. Furthermore, considering the deep dependency of EdgeText on text edges, a bilateral enhanced perception (BEP) module is designed. It encourages our model to pay attention to the recognition of edge features. Additionally, to accelerate the learning of the curve function parameters, we introduce a proportional integral loss (PI-loss) to force the proposed model to focus on the curve distribution and avoid being disturbed by text scales. Ablation experiments demonstrate that EdgeText can fit scene texts compactly and naturally. Comparisons show that EdgeText is superior to existing methods on multiple public datasets.

\end{abstract}

\begin{IEEEkeywords}
Scene text detection, text shape representation, edge approximation, parameterized modeling
\end{IEEEkeywords}

\section{Introduction}
\label{Introduction}
\IEEEPARstart{S}{cene} text detector~\cite{zhang2023video,guan2022industrial,cao2021all}, a front component of the text spotting pipeline, is responsible for determining text locations and feeding for the following components to recognize texts. The design of an efficient text representation helps formulate compact instance contours to irregular scene texts with fewer procedures, which encourages the detection module to avoid bringing the noise into the recognition stage and optimizes the detection pipeline from complex contour-rebuilding processes. Therefore, \textbf{\textit{how to represent irregular scene texts efficiently}} is important for text detection models. 

\begin{figure}
	\centering
	\includegraphics[width=.47\textwidth]{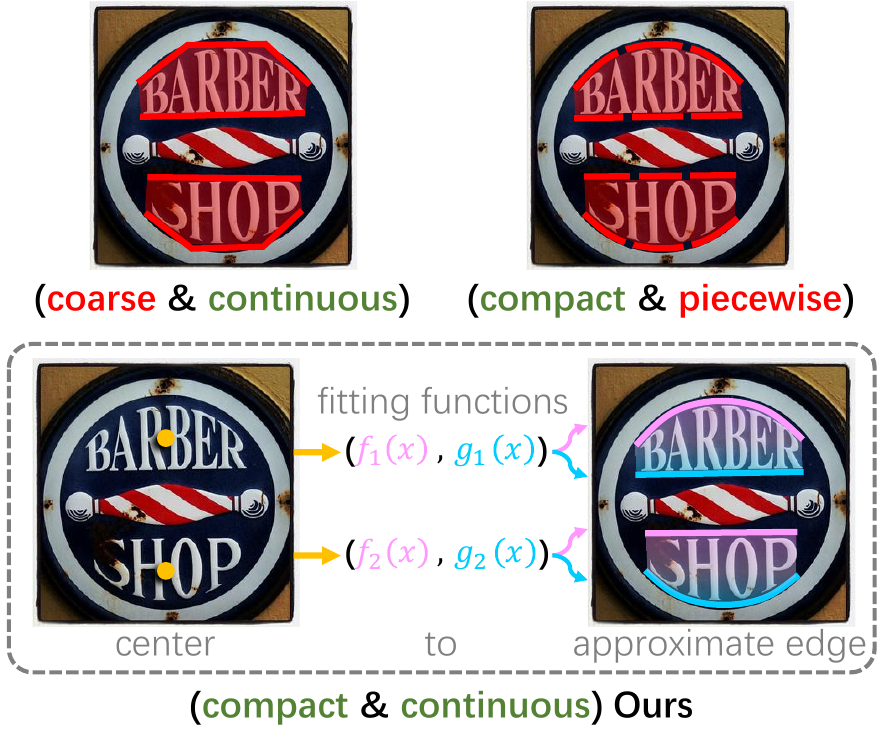}
	\caption{Visualization of the differences between our edge approximation text representation method (bottom sub-figure) and the current popular representations of box-to-polygon strategy (left top sub-figure) and piecewise fitting process (right top sub-figure).}
	\label{V1}
\end{figure}

Current popular practices are to represent irregular text shapes via box-to-polygon strategy~\cite{dai2021progressive} or decompose a contour into pieces for fitting gradually~\cite{ye2023deepsolo,liu2021abcnet,long2018textsnake}. Specifically, the former strategy starts from rectangular box detection. It then samples a few points along two long sides of boxes and maps these points to text region boundaries. This strategy ensures intuitive polygon generation processes of irregular texts, but hard to fit compactly (such as the left top of Fig.~\ref{V1}), which may pass noises to the following components and make the recognition task difficult. Different from the box-to-polygon strategy, existing piecewise fitting methods are able to represent scene texts with compactness boundaries (the sub-figure at the right top of Fig.~\ref{V1}), but the gradual contour reconstruction process complicates the whole inference pipeline and the framework training optimization.

Following the above issues, we aim to represent scene texts compactly and continuously to avoid the limitations existing in the box-to-polygon and piecewise fitting methods. Concretely, it is observed that the text is a string with a series of characters (such as English letters, Arabic numerals, and Chinese, etc), which makes the scene text contour always can be seen as a ribbon-like shape that consists of two curves along the text's long edges. Therefore, the scene text contour can be reconstructed easily once the curves are determined. Based on the above observation, we formulate the irregular text representation problem as an approximation process of parameterized curves toward text edges, and an end-to-end text detector framework, namely \textit{EdgeText}, is constructed following the design of edge approximation text representation. In the inference stage (shown in Fig.~\ref{V2}), EdgeText starts with locating text centers based on the pixel-level foreground classification confidence map, then creating text edge approximation curves predefined as mathematical formula parameters by using the center's corresponding visual information. Meanwhile, determining truncation points defined in the image pixel coordinate system from text location features. Next, truncating the approximation curves with the points for extracting curve segments to reconstruct irregular text contours. As described before, the designed curve box relies on the curve fitting functions of approximate edges and truncation points deeply. To pursue accurate curve boxes in the reconstruction process of the inference stage, a bilateral enhanced perception (BEP) module is introduced. The module encourages EdgeText to focus on the two long truncated edges that are created from approximate edges, which helps our model improve the recognition of the corresponding edge features. Additionally, a proportional integral loss (PI-loss) is proposed. It treats the integral of the scaled difference between two curves as the optimization objective, which avoids the preference problem toward large-scale instances and helps improve the parameter learning of the curve fitting functions. The main contributions of this paper are as follows:

\begin{figure}
	\centering
	\includegraphics[width=.48\textwidth]{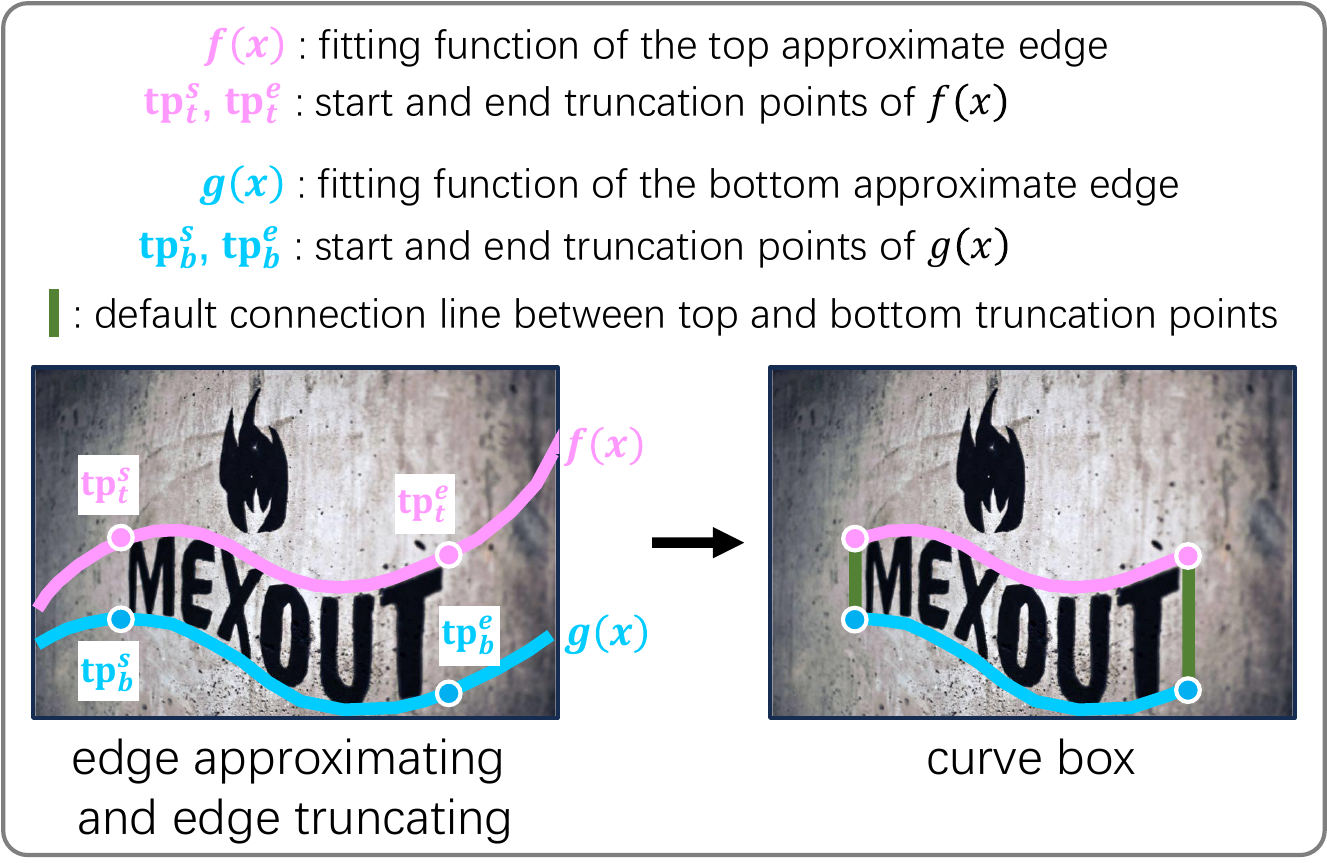}
	\caption{Detail visualization of the proposed edge approximation text representation method. It fits texts with the curve box that is constructed based on the approximate edge and truncation points. The same as the traditional bounding box representation, our \textbf{curve box} enjoys a simple reconstruction process. Especially, it can fit irregular texts accurately instead of covering rectangular shapes only like the traditional bounding box.}
	\label{V2}
\end{figure}

\begin{enumerate}
	\item An edge approximation representation method is proposed following the scene text characteristics of ribbon-like shapes. It represents arbitrary-shaped texts via parameterized curve fitting functions and truncation points, which ensures compact and continuous rebuilt contours to achieve accurate text representation efficiently.
	
	\item A bilateral enhanced perception (BEP) module is designed to improve the feature recognition ability toward text edges. It helps the model accurately perceive text edge features for generating reliable curve boxes.
	
	\item A proportional integral loss (PI-loss) is introduced. It treats the integral of the scaled difference between two curves as the optimization objective, which helps the model focus on the curve distribution and avoid being disturbed by text scales.
	
	\item An efficient text detection framework (EdgeText) is constructed using the proposed edge approximation representation method, BEP, and PI-loss. It can fit arbitrary-shaped scene texts compactly with a simple and intuitive practice, which alleviates the limitations that exist in current popular models.
\end{enumerate}

\section{Related Work}
\label{Related Work}
The strong feature expression ability of deep learning networks progresses scene text detection models in the aspect of feature understanding of the complex background and text. To fit varied shapes of scene texts, existing methods focus on fitting text contours as accurately as possible. They can be divided into regular and irregular methods according to text shapes. In this section, we will introduce these methods briefly.

\subsection{Methods toward the Regular Text Shape}
Regular-shaped texts denote the instances that can be fitted by rectangle and quadrilateral boxes accurately. Early text detectors~\cite{liu2017deep,ma2018arbitrary} are inspired by traditional two-stage object detection framework~\cite{ren2015faster} to represent texts via anchor-to-box strategy. These methods start with extracting the text region of interest (RoI) based on predefined anchors and then refining the region vertex locations for generating regular text boxes. Especially, Liu~$et~al.$~\cite{liu2017deep} and Ma~$et~al.$~\cite{ma2018arbitrary} introduced rotation anchors for fitting multi-oriented texts. Though they achieve comparable performance, the two-stage framework makes the pipeline complex. With the proposing of Single Shot MultiBox Detector (SSD)~\cite{liu2016ssd}, one-stage text detection methods~\cite{wang2021towards,liao2018textboxes++,tian2017wetext} inherit the corresponding design for alleviating the problem of the complex inference process. Different from the previous methods, the authors~\cite{shi2017detecting,liao2017textboxes,liao2018rotation} abandoned the RoI refining process in the two-stage framework and regressed the vertex offsets between anchors and text boxes directly, which optimizes the overall detection pipeline effectively. Considering the anchor mechanism results in model performance deeply relies on prior knowledge of dataset statistics, researchers followed the anchor-free works~\cite{redmon2016you,huang2015densebox,tian2019fcos} for avoiding the prior dependency problem that exists in those methods while further simplifying the whole detection process. Specifically, the anchor-free text detector~\cite{zhou2017east} executes dense prediction toward the offset between the current pixel location and box vertices, which enjoys better generalization performance compared with previous methods. The above models achieve comparable detection results on public regular text detection datasets. However, with the appearance of irregular-shaped scene texts, these methods are hard to fit accurately.

\begin{figure*}
	\centering
	\includegraphics[width=.99\textwidth]{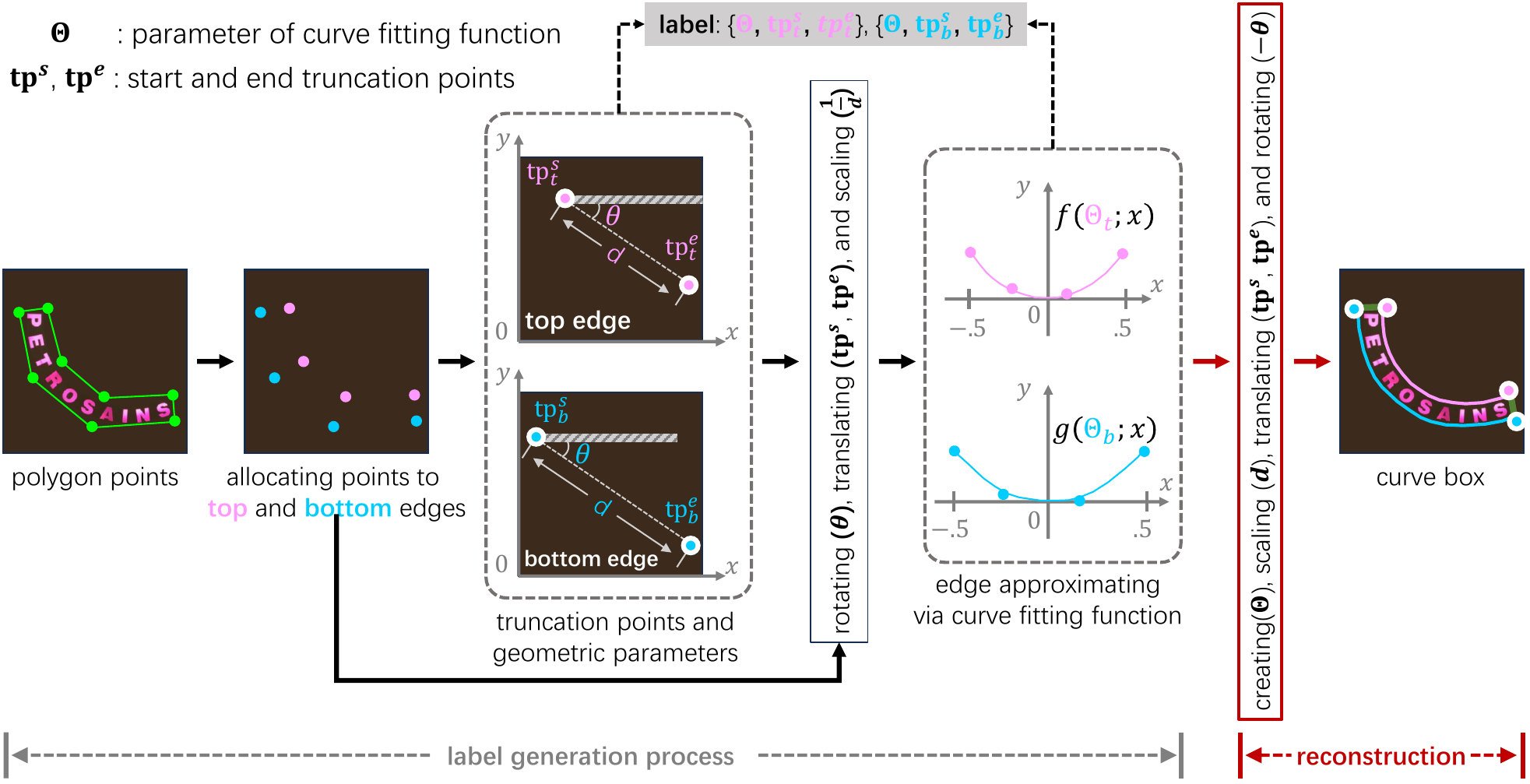}
	\caption{Visualization of the processes of the label generation and curve box reconstruction. $f(\mathrm{\Theta}_t;x)$ and $g(\mathrm{\Theta}_b;x)$ are the same curve-fitting function except for different parameters. Polynomial is adopted as the curve fitting function in this paper.}
	\label{V3}
\end{figure*}

\subsection{Methods toward the Irregular Text Shape}
A valid representation design for irregular-shaped texts can help avoid feeding interference information to the text recognizer for improving the accuracy of recognition results. Initially, related methods~\cite{zhang2020deep,deng2023progressive,wang2019arbitrary} fit irregular texts based on the bounding box representation. Concretely, Zhang~$et~al.$~\cite{zhang2020deep} decomposed the text into a series of characters and detected them via traditional object detection methods. The final text contours can be obtained by linking character boxes. Some other authors~\cite{zhang2019look,wang2020contournet,wang2020all} started with detecting the whole text box. Then, they sampled a few key points along the long sides of the box. Next, those points were mapped to text boundaries for fitting. In the end, text contours can be reconstructed by connecting mapped key points in a clockwise or counterclockwise direction. Besides the above detection-based representation design, an intuitive idea is to segment text regions~\cite{liao2020mask,tian2019learning,xu2021rethinking} by using the strong pixel-level detection ability of the image segmentation techniques~\cite{long2015fully,ronneberger2015u}. However, the text adhesion or occlusion phenomenon always exists in natural scenes, it will lead to detect multiple instances as one if segmenting text regions directly. Considering the above issues, researchers reconstructed the segmentation-based detection pipeline~\cite{xu2019textfield,liao2020real,zhao2024cbnet,dai2019deep,zhang2022arbitrary}. Specially, Baek~\cite{baek2019character} adopted a character-connection strategy. The author segmented every char and grouped them together that belong to the same text for rebuilding text contours. Different from Baek~\cite{baek2019character}, Zhu~\cite{zhu2021fourier} and Liu~\cite{liu2020abcnet} represented text contour via mathematical fitting functions (such as Fourier series and Bessel curve). However, the complex transform process from the corresponding fitting function to the text contour makes it hard to optimize the model and achieve text rebuilding via fewer steps. Current methods either are hard to cover irregular text regions compactly or rely on complex fitting processes, how to represent scene texts in an efficient manner is still under exploration.

\section{Methodology}
\label{Methodology}
In this section, we introduce the designed edge approximation-based text representation method first. Then, the overall constructed detection framework (EdgeText) is described combined with the pipeline figure. Next, the bilateral enhanced perception (BEP) module is visualized to illustrate the corresponding structure in detail. In the end, loss functions, especially the proposed proportional integral loss, are given to explain the model optimization process.

\subsection{Edge Approximation Representation}
\label{Edge Approximation Representation}
Designing an efficient representation method for irregular-shaped scene texts can help cover text regions compactly with fewer procedures, which ensures reliable location information feeding for the following recognition task while keeping a simple inference pipeline. Considering current popular practices either are hard to cover texts compactly or rely on complex fitting processes, we proposed an edge approximation-based text representation method in this paper.

As mentioned in Fig.~\ref{V2} before, the proposed representation method fits irregular texts via the curve box created through the two processes of edge approximating and edge truncating. As shown in Fig.~\ref{V3}, given a text instance labeled by a series of polygon points $\mathrm{P}=[p_1, p_2, ..., p_{2k}]$, where $p_k=\begin{bmatrix}x_k\\y_k\end{bmatrix}$, which is the $k$-th point location in the horizontal pixel coordinate system. $2k$ denotes the number of points. We first define the two long sides of the text as top and bottom edges respectively and allocate those polygon points to the two edges equally. The top edge points $\mathrm{P}_t=[p_1, p_2, ..., p_{k}]$ and the bottom edge points $\mathrm{P}_b=[p_{k+1}, p_{k+2}, ..., p_{2k}]$ can be obtained. Then, the parameter $\mathrm{\Theta}$ of the curve fitting function, and start and end truncation points ($\mathrm{tp}^s$ and $\mathrm{tp}^e$) are generated as the training labels, which are used for reconstructing text contours. \textbf{Here we take the top edge as an example to explain the generation process, and the label generation process of the bottom edge is the same as the top edge}. Concretely, the points $p_1$ and $p_k$ in $\mathrm{P}_t$ are picked as the truncation points ($\mathrm{tp}^s_t$ and $\mathrm{tp}^e_t$) of the top edge. And $\mathrm{P}_t$ are rotated, translated, and scaled according to truncation points for generating a series of transformed points $\mathrm{P}^{'}_t$ of horizontal, symmetrical to the origin of the coordinate system, and the scaling to -0.5 to 0.5 on X-axis, which helps the model deviate from the interference of spatial distribution and focus on fitting curve shape of $\mathrm{P}_t$. The rotating, translating, and scaling processes can be formulated by mathematical equations as follows:
\begin{eqnarray} 
\label{e1}
\begin{gathered}
	\mathrm{P}^R_t=
	\mathrm{R}_t(\mathrm{P}_t-\mathrm{tp}^s_t)
	=
	\begin{bmatrix}
		\cos \theta_t & -\sin \theta_t \\
		\sin \theta_t & \cos \theta_t
	\end{bmatrix}(\mathrm{P}_t-\mathrm{tp}^s_t),\\
	\mathrm{P}^T_t=\mathrm{P}^R_t-\frac{1}{2}(\mathrm{P}^R_{t,\frac{k-1}{2}}+\mathrm{P}^R_{t,\frac{k}{2}}),\\
	\mathrm{P}'_t=\mathrm{P}^S_t=\frac{1}{d_t}\mathrm{P}^T_t,
\end{gathered}
\end{eqnarray} 
where $\mathrm{P}^R_t$, $\mathrm{P}^T_t$, and $\mathrm{P}^S_t$ indicate the points that are rotated, translated, and scaled based on $\mathrm{P}_t$. $\mathrm{R}_t$ is the rotation matrix. $\theta_t$ denotes the angle between the vector $\overrightarrow{\mathrm{tp}^s_t\mathrm{tp}^e_t}$ and X-axis. $\mathrm{P}^R_{t,\frac{k-1}{2}}$ and $\mathrm{P}^R_{t,\frac{k}{2}}$ are $\frac{k-1}{2}$-th and $\frac{k}{2}$-th points in $\mathrm{P}^R_t$. $d_t$ is the Euclidean distance between $\mathrm{tp}^s_t$ and $\mathrm{tp}^e_t$.  After transforming polygon points from $\mathrm{P}_t$ to $\mathrm{P}^{'}_t$ through the three operations in Equation~\ref{e1}, the corresponding curve fitting function $f(\mathrm{\Theta}_t;x)$ that approximating text edge can be formulated, where $\Theta_t$ is the parameter of curve fitting function that the model need to learn. For the text bottom edge, the corresponding truncation points $\mathrm{tp}^s_b$ and $\mathrm{tp}^e_b$, transformed polygon points $\mathrm{P}^{'}_b$ and curve fitting function $g(\mathrm{\Theta}_b;x)$ can be obtained in the same way. \textbf{Notably}, $f(\mathrm{\Theta}_t;x)$ and $g(\mathrm{\Theta}_b;x)$ are the same curve-fitting function except for different parameters and the polynomial is adopted as the curve-fitting function in this paper:
\begin{eqnarray} 
	\label{e2}
	\begin{gathered}
		f(\mathrm{\Theta}_t;x)=g(\mathrm{\Theta}_b;x)= {\textstyle \sum_{i=1}^{m}} \Theta_{i}x^i+c,\\
		\mathrm{\Theta}_t = \{\mathrm{\Theta}_{i,t},c|i=1,2,...,m\},\\
		\mathrm{\Theta}_b = \{\mathrm{\Theta}_{i,b},c|i=1,2,...,m\},
	\end{gathered}
\end{eqnarray} 
where $m$ denotes the highest degree of the polynomial and $c$ represents the constant term.

\begin{figure}
	\centering
	\includegraphics[width=.48\textwidth]{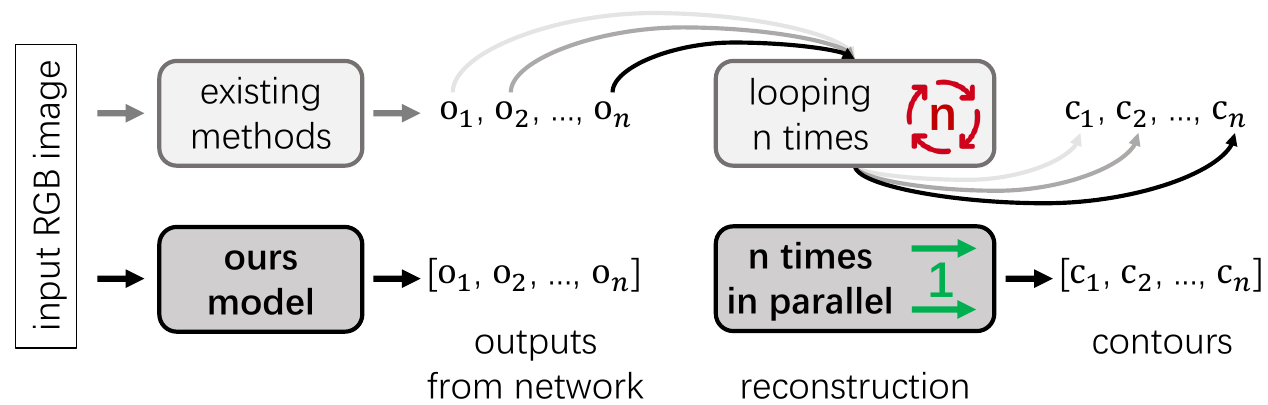}
	\caption{Visualization of the contour reconstruction differences between existing methods and ours EdgeText. Compared with existing box-to-polygon strategy-based or piecewise fitting methods that have to reconstruct every single text one by one until all instance contours are generated, EdgeText rebuilds all text curve boxes in the input image in parallel, which enjoys a more intuitive contour rebuilding process with fewer procedures.}
	\label{V4}
\end{figure}

In the inference stage, benefiting from the advantages brought by the edge approximation-based representation method and the above transformation operations in Equation~\ref{e1}, the reconstruction process of text contours with different rotated angles, spatial locations, and scales can be treated as a matrix manipulation about curve fitting function sampling points, which allows rebuilding all text contours in the input image in parallel. Compared with existing box-to-polygon strategy-based or piecewise fitting methods that have to reconstruct every single text one by one until all instance contours are generated, EdgeText enjoys a more intuitive contour rebuilding process with fewer procedures (as shown in Fig.~\ref{V4}). Specifically, with the predicted parameters $\tilde{\mathrm{\Theta}}_t$ and $\tilde{\mathrm{\Theta}}_b$ of the curve fitting functions $f(\tilde{{\mathrm{\Theta}}}_t;x)$ and $g(\tilde{\mathrm{\Theta}}_b;x)$, and start and end truncation points, the transformed dense curve points $\tilde{\mathrm{P}}^{'}$ can be obtained by following equations:
\begin{eqnarray} 
	\label{e3}
	\begin{gathered}
		\tilde{\mathrm{P}}^{'}=\begin{bmatrix}
			\tilde{\mathrm{P}}^{'}_1 \\
			\tilde{\mathrm{P}}^{'}_2\\
			\vdots \\
			\tilde{\mathrm{P}}^{'}_n
		\end{bmatrix},~~
			\tilde{\mathrm{P}}^{'}_n
		=\begin{bmatrix}
			\tilde{\mathrm{P}}^{'}_{n,t}\\
			\tilde{\mathrm{P}}^{'}_{n,b}
		\end{bmatrix}
		=\begin{bmatrix}
			\begin{bmatrix}
				X\\f(\tilde{\mathrm{\Theta}}_{n,t};X)
			\end{bmatrix} \\
			\begin{bmatrix}
				X\\g(\tilde{\mathrm{\Theta}}_{n,b};X)
			\end{bmatrix}
		\end{bmatrix},\\
		X=[x_1,x_2,...,x_n], -0.5\le x\le0.5, n>k,
	\end{gathered}
\end{eqnarray} 
where $n$ denotes the number of text instances in the input image, $ (\cdot)_{n,t}$ and $(\cdot)_{n,b}$ are the predicted values of the top and bottom edges of the $n$-th text instance respectively.

\begin{figure*}
	\centering
	\includegraphics[width=.99\textwidth]{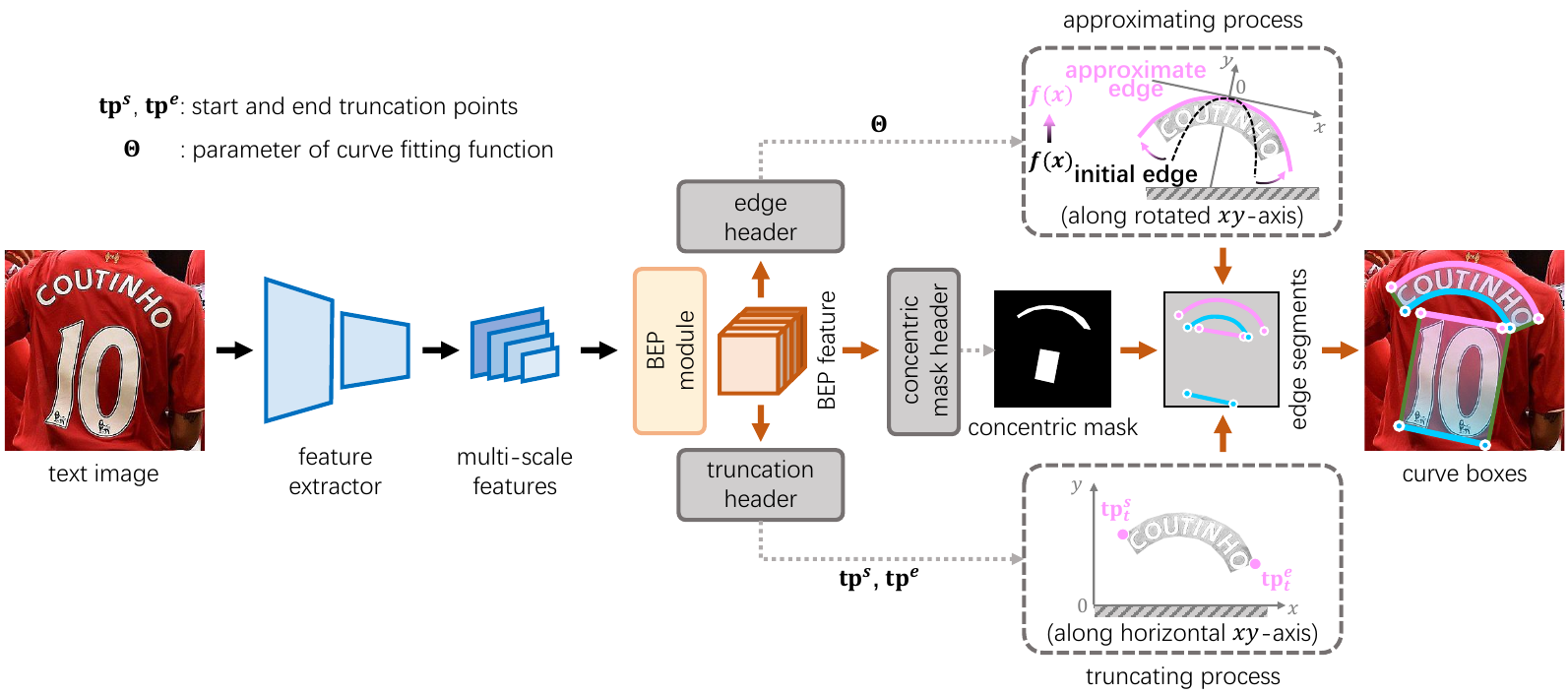}
	\caption{Pipeline visualization of the constructed EdgeText, which is composed of a feature extractor, BEP module, and three headers of edge, truncation, and concentric mask. These headers are responsible for the parameter prediction of edge approximating curve fitting function, the determination of truncation points, and the location of text instances.}
	\label{V5}
\end{figure*}

With the transformed dense curve points $\tilde{\mathrm{P}}^{'}$, the final text contours can be generated by scaling, translating, and rotating operations, which enjoys inverse operation orders with label generation in Equation~\ref{e1}. The corresponding  scaling process can be formulated as follows:
\begin{eqnarray} 
	\label{e4}
	\begin{gathered}
		\tilde{\mathrm{P}}^{'S}=\begin{bmatrix}
			\tilde{d}_1 \\
			\tilde{d}_2\\
			\vdots \\
			\tilde{d}_n
		\end{bmatrix}\odot \tilde{\mathrm{P}}^{'},~~
		\tilde{d}_n
		=\begin{bmatrix}
			\tilde{d}_{n,t} \\
			\tilde{d}_{n,b}
		\end{bmatrix},
	\end{gathered}
\end{eqnarray} 
where $\tilde{d}$ is the Euclidean distance computed by the predicted truncation points. With the scaling point matrix $\tilde{\mathrm{P}}^{'S}$, translating point matrix $\tilde{\mathrm{P}}^{'T}$ can be obtained by:
\begin{eqnarray} 
	\label{e5}
	\begin{gathered}
		\tilde{\mathrm{P}}^{'T}=\tilde{\mathrm{P}}^{'S}+( \begin{bmatrix}
			\tilde{\mathrm{tp}}^s_1 \\
			\tilde{\mathrm{tp}}^s_2 \\
			\vdots \\
			\tilde{\mathrm{tp}}^s_n
		\end{bmatrix}-\tilde{\mathrm{P}}^{'S}_{:,0}),~~
		\tilde{\mathrm{tp}}^s_n
		=\begin{bmatrix}
			\tilde{\mathrm{tp}}^s_{n,t} \\
			\tilde{\mathrm{tp}}^s_{n,b}
		\end{bmatrix},
	\end{gathered}
\end{eqnarray} 
where $\tilde{\mathrm{tp}}^s$ is the predicted start truncation point. Based on the above $\tilde{\mathrm{P}}^{'T}$, the final point matrix $\tilde{\mathrm{P}}$ used for reconstructing text contours is computed by:
\begin{eqnarray} 
	\label{e6}
	\begin{gathered}
		\tilde{\mathrm{P}}=\tilde{\mathrm{P}}^{'R}=
		\begin{bmatrix}
			\tilde{\mathrm{R}}_1 \\
			\tilde{\mathrm{R}}_2 \\
			\vdots \\
			\tilde{\mathrm{R}}_n
		\end{bmatrix}\odot\tilde{\mathrm{P}}^{'T},~~
		\tilde{\mathrm{R}}_n=
		\begin{bmatrix}
			 \tilde{\mathrm{R}}_{n,t} \\
			  \tilde{\mathrm{R}}_{n,b}
		\end{bmatrix},
	\end{gathered}
\end{eqnarray} 
where $\tilde{\mathrm{R}}$ is a rotation matrix that can be referred the $\mathrm{R}_t$ in Equation~\ref{e1}, where the $\theta$ is obtained by the predicted truncation points ($\tilde{\mathrm{tp}}^s$ and $\tilde{\mathrm{tp}}^e$) and X-axis.

\subsection{Overall Framework}
\label{Overall Framework}
As introduced in the above section, the edge approximation-based representation method helps fit irregular texts compactly in an intuitive way, which encourages avoiding the limitations in current popular practices. Following the designed text representation method, we construct a novel detection framework (EdgeText), and the structure is described in this section.

The framework comprises a feature extractor, BEP module, and three headers that of edge, truncation, and concentric mask (as shown in Fig.~\ref{V5}). Feature extractor includes traditional backbone~\cite{simonyan2014very,he2016deep,xie2017aggregated} and feature pyramid network (FPN)~\cite{lin2017feature}, which is responsible to provide multi-scale visual features $\mathrm{F}_1$, $\mathrm{F}_2$, $\mathrm{F}_3$, and $\mathrm{F}_4$ corresponding to the image size of $\frac{1}{4}$, $\frac{1}{8}$, $\frac{1}{16}$, and $\frac{1}{32}$ respectively for the following prediction headers. Considering the recognition of text edge features is important for rebuilding reliable curve boxes in the inference stage, we design a bilateral enhanced perception (BEP) module. It takes the multi-scale visual features as input and optimizes the $\mathrm{F}_1$ with a Gaussian truncated edge segment, which encourages EdgeText to recognize edge shape distribution more accurately. The BEP final outputs a BEP feature $\mathrm{F}_{bep}\in \mathbb{R}^{\frac{H}{4},\frac{W}{4},ch}$ concatenated via enhanced multi-scale visual features, where $H$ and $W$ are the height and width of input image respectively. $ch$ denotes the number of the feature channels. After the feature enhancing process, the concentric mask~\cite{DBLP:journals/tip/YangCXYW22} (a shrink-mask used for locating text instances without the influence of occlusion or adhesion problem) header then segments the text concentric mask to determine the corresponding text location. In the end, edge and truncation headers then predict the parameter $\Theta$ of the curve fitting function and start and end truncation points ($\mathrm{tp}^s$ and $\mathrm{tp}^e$) (referred to Section~\ref{Edge Approximation Representation}) based on the BEP feature and text location information brought by concentric masks. Specifically, The three headers ($\mathrm{h}_{conc}$, $\mathrm{h}_{edge}$, and $\mathrm{h}_{trun}$) consists of a stack of fully convolution layers, the structure can be expressed as follows:
\begin{eqnarray} 
	\label{e7}
	\begin{gathered}
		\mathrm{h}_{conc}=\mathrm{C}^{ch_c}_3(\mathrm{C}^{\frac{ch}{2}}_3(\mathrm{C}^{ch}_3(\mathrm{C}^{ch}_1(\mathrm{F}_{bep})))),\\
		\mathrm{h}_{edge}=\mathrm{C}^{ch_e}_3(\mathrm{C}^{\frac{ch}{2}}_3(\mathrm{C}^{ch}_3(\mathrm{C}^{ch}_1(\mathrm{F}_{bep})))),\\
		\mathrm{h}_{trun}=\mathrm{C}^{ch_t}_3(\mathrm{C}^{\frac{ch}{2}}_3(\mathrm{C}^{ch}_3(\mathrm{C}^{ch}_1(\mathrm{F}_{bep})))),
	\end{gathered}
\end{eqnarray} 
where $\mathrm{C}^{ch}_1$ denotes the convolution layer with 1$\times$1 kernel and outputs the features with $ch$ channels.  $ch_c$, $ch_e$, and $ch_t$ are the numbers of the output channels of the three headers respectively. $ch_c$ is set to 1 for representing the text category. $ch_e$ is the number of curve fitting function parameters (e.g., it is set as 2, 3, or 4 when the polynomial's degrees are 1,2 or 3). $ch_t$ is set to 8 for encoding the coordinates $(x, y)$ of four truncation points ($\mathrm{tp}^s_t$, $\mathrm{tp}^e_t$, $\mathrm{tp}^s_b$, and $\mathrm{tp}^e_b$). 

For each text, the values on the concentric mask center locations of $\mathrm{h}_{edge}$ are picked as predicted curve fitting function parameters. Different from the parameters, considering the large offsets between center points and truncation points, we choose the lateral vein prediction strategy proposed in LeafText~\cite{yang2023text} to determine the truncation point locations. The strategy extracts text center lines from concentric masks at first. Then, the start and end points of the center lines are determined. Finally, the start truncation points ($\mathrm{tp}^s_t$ and $\mathrm{tp}^s_b$) of the text's top and bottom edges are obtained by the values from $\mathrm{h}_{trun}$ on the center line start point locations.  For the end truncation points ($\mathrm{tp}^e_t$ and $\mathrm{tp}^e_b$), they are determined by the values from $\mathrm{h}_{trun}$ on the center line endpoint locations. The above strategy allows us to predict shorter offsets between the start and end points of the center lines and truncation points instead of regressing the larger offsets between center points and truncation points directly, which helps to locate truncation points more accurately. With the parameters and truncation points, the curve boxes can be reconstructed by the process introduced in the above section.

\subsection{Bilateral Enhanced Perception Module}
To improve EdgeText's recognition ability of text edge features, we design a bilateral enhanced perception (BEP) module. It optimizes the network via Gaussian truncated edge segment supervision labels, encouraging EdgeText to recognize edge distribution more accurately. The module structure will be illustrated in this section.

\begin{figure}
	\centering
	\includegraphics[width=.48\textwidth]{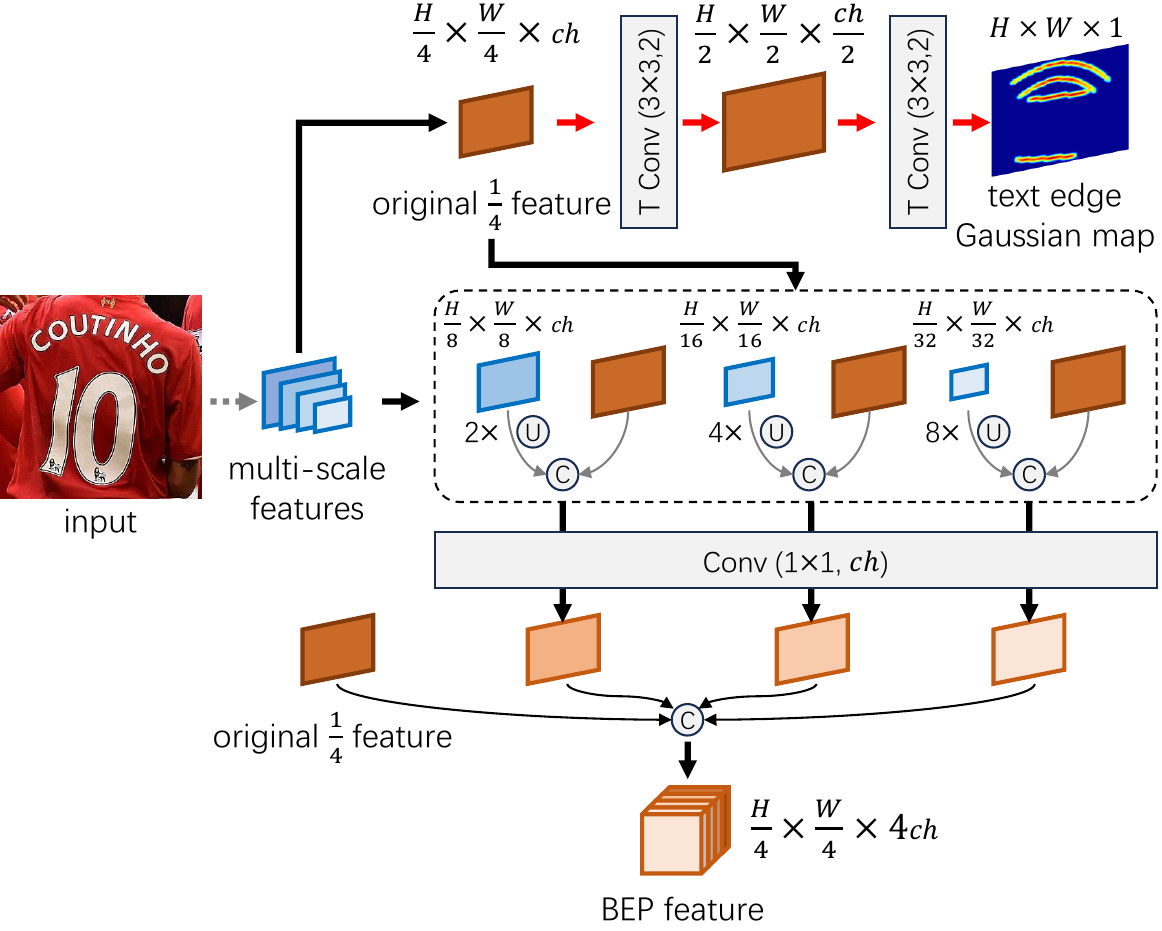}
	\caption{Visualization of the proposed BEP module. It takes the multi-scale features from the feature extractor as input for generating a BEP feature to feed the following prediction headers.}
	\label{V6}
\end{figure}

As visualized in Fig.~\ref{V6}, the BEP module takes the multi-scale features from the feature extractor (referred to Fig.~\ref{V5}) as input for generating a bilateral enhanced perception (BEP) feature to feed the following prediction headers. Concretely, the module up-samples $\mathrm{F}_1$ to the size of $H\times W$ via two transposed convolution layers to predict the text edge confidence map first. It enhances the fine-grained feature ($\mathrm{F}_1$) in multi-scale features to recognize the text edge shapes and the corresponding truncation point locations. Then, $\mathrm{F}_2$, $\mathrm{F}_3$, and $\mathrm{F}_4$ are up-sampled to the size of $\mathrm{F}_1$ for generating three concatenate features. Next, the channels of these features are reduced to $ch$ via a convolution layer with $1\times 1$ kernel. In the end, the above three concatenate features with $ch$ channels are further concatenated with the enhanced $\mathrm{F}_1$ to generate the final BEP feature. The concatenate operation between fine-grained feature $\mathrm{F}_1$ and coarse-grained features $\mathrm{F}_2$, $\mathrm{F}_3$, and $\mathrm{F}_4$ helps the BEP feature enjoys the comprehensive ability to recognize curve box related features.

\begin{figure*}
	\centering
	\includegraphics[width=.99\textwidth]{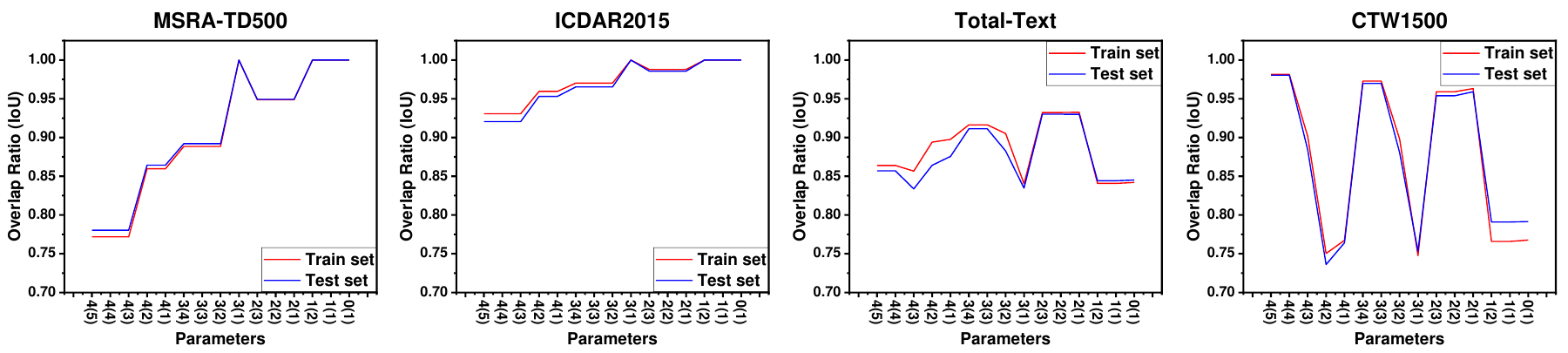}
	\caption{Analysis of the overlap ratio between reconstructed curve boxes based on edge approximation representation and ground-truth contours via the metric of Intersection of Union (IoU) under different settings of fitting function parameters. The tick label on the Parameters-axis illustrates the specific parameter setting (e.g., 4(3) means the highest degree $m$ of polynomial is set to 4. Only the parameters of $\{\Theta_i,|i=2,3,4\}$ are adopted to fitting the text edge without the parameters of $\{\Theta_i,c|i=1\}$).}
	\label{V7}
\end{figure*}

\subsection{Loss Function}
EdgeText represents irregular texts via the curve box that consists of the curve fitting function and truncation points. As described in Fig.~\ref{V5}, the edge header and truncation header are responsible for predicting the parameters and points respectively, and the BEP module helps our model to recognize the features of edge shape and truncation points. To optimize the proposed framework effectively, we construct a comprehensive loss function $\mathcal{L}$, which comprises edge header loss $\mathcal{L}_{edge}$, truncation header loss $\mathcal{L}_{trun}$, and BEP module loss $\mathcal{L}_{bep}$. The comprehensive loss function $\mathcal{L}$ can be expressed as follows:
\begin{eqnarray} 
	\label{e8}
	\begin{gathered}
		\mathcal{L}=\alpha\mathcal{L}_{edge}+\beta\mathcal{L}_{trun}+\gamma\mathcal{L}_{bep},
	\end{gathered}
\end{eqnarray} 
where $\alpha$, $\beta$, and $\gamma$ are the importance coefficients of $\mathcal{L}_{edge}$, $\mathcal{L}_{trun}$, and $\mathcal{L}_{bep}$, respectively. They are set to 0.5, 0.5, and 1.0 in the following experiments.

For $\mathcal{L}_{edge}$, considering the varied scales of scene texts, we propose a proportional integral loss $\mathcal{L}_{pi}$ to optimize the edge header. $\mathcal{L}_{pi}$ creates curves according to the predicted parameter $\tilde{\mathrm{\Theta}}$ and the corresponding ground-truth ${\mathrm{\Theta}}$ in the range from -0.5 to 0.5 on X-axis. Then, the integral of the absolute difference between two curves in this range is taken as the optimization objective. Here, we take the top edge as an example to formulate the computation process:
\begin{eqnarray} 
	\label{e9}
	\begin{gathered}
		\mathcal{L}_{edge}=\mathcal{L}_{pi}=\int_{-0.5}^{0.5}|f(\Theta;x)-f(\tilde{\Theta} ;x)|dx,
	\end{gathered}
\end{eqnarray} 
where $\Theta$ is the parameter obtained from the curve fitting function of transformed polygon points instead of the original label points (as explained in Fig.~\ref{V3}). Therefore, EdgeText can learn the overall shape features of the text edge in the range of -0.5 to 0.5 on the X-axis and not be affected by text scales, which ensures our model balances the importance of different instances with varied scales and accelerates the optimization compared with computing loss according to original points directly. Considering the image is pixel-level matrix and the independent variable $x$ is not continuous, we transform the Equation~\ref{e9} to the discrete form:
\begin{eqnarray} 
	\label{e10}
	\begin{gathered}
		\mathcal{L}_{pi}=\lim_{N \to \infty} {\textstyle \sum_{i=1}^{N}}|f(\Theta;x_i)-f(\tilde{\Theta} ;x_i)|\Delta x_i,\\\Delta x=\frac{1}{N},~~x\in [-0.5,0.5],
	\end{gathered}
\end{eqnarray} 
where $N$ denotes the number of sample points in the range of -0.5 to 0.5 on the X-axis and when $N$ approaching infinity, the above discrete form is equal to continuous form. To compute the loss in the training stage, $N$ is set as a finitude number in this paper empirically and the Equation~\ref{e10} can be further simplified as following expression:
\begin{eqnarray} 
 	\label{e11}
 	\begin{gathered}
 		\mathcal{L}_{pi}={\textstyle \sum_{i=1}^{N}}|f(\Theta;x_i)-f(\tilde{\Theta} ;x_i)|,\\x_i=-0.5+\frac{i}{N}.
 	\end{gathered}
 \end{eqnarray} 
 
For truncation header loss $\mathcal{L}_{trun}$ and BEP module loss $\mathcal{L}_{bep}$, they are responsible for supervising the distance regression and semantic segmentation tasks respectively. Considering smooth-$l_1$ loss~\cite{ren2015faster} is more robust to outliers and less susceptible to extreme values than $l_2$ loss, thus we adopt smooth-$l_1$ loss as $\mathcal{L}_{trun}$ for optimization of the truncation header in the training stage for ensuring model's stability and generalization ability. Furthermore, considering the sample imbalance between the background and the text edge region, Dice loss~\cite{milletari2016v} is introduced to evaluate the BEP module loss $\mathcal{L}_{bep}$, which can be expressed as the following equation:
\begin{eqnarray} 
	\label{e12}
	\begin{gathered}
		\mathcal{L}_{bep}=1-\frac{2\times |\mathrm{M}_{pred}\cap \mathrm{M}_{gt}|+\varepsilon }{|\mathrm{M}_{pred}|+|\mathrm{M}_{gt}|+\varepsilon },
	\end{gathered}
\end{eqnarray} 
where $\mathrm{M}_{pred}$ and $\mathrm{M}_{gt}$ are the predicted text edge and the corresponding ground-truth. $\varepsilon$ is used to avoid the situation where there may be no positive samples in $\mathrm{M}_{gt}$, which is set to 1 in the following experiments empirically.

\begin{figure*}
	\centering
	\includegraphics[width=.95\textwidth]{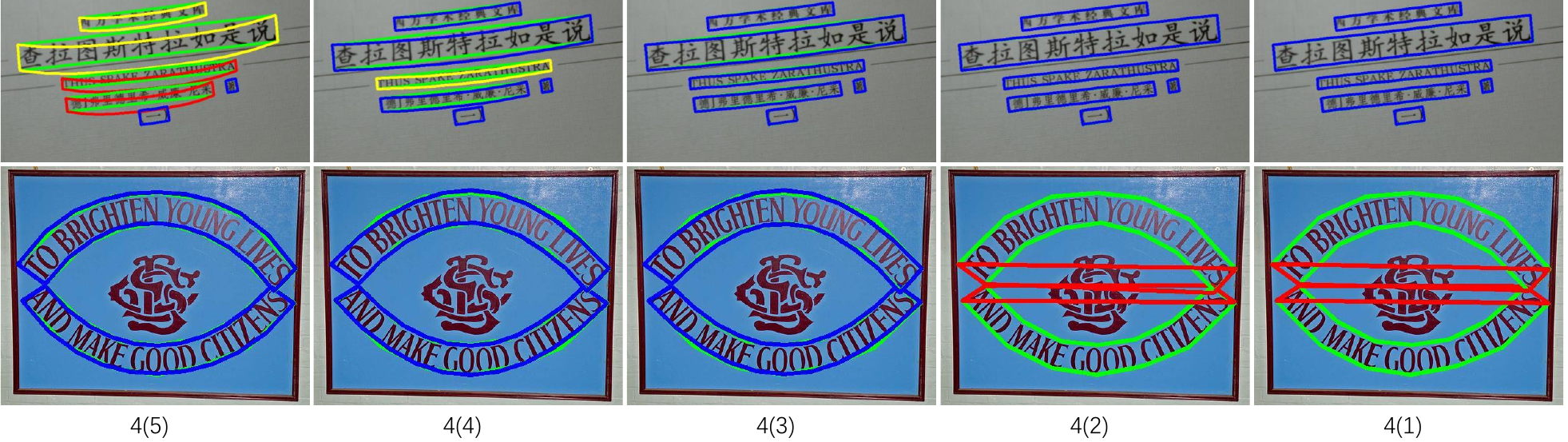}
	\caption{Visualization of the fitting results generated by the proposed edge approximation representation method toward regular and irregular-shapes texts. Green contours are the ground-truth labels.  \textbf{Blue}, \textbf{yellow}, and \textbf{red} contours are the generated high-quality, medium-quality, and low-quality curve boxes respectively, where different level qualities are defined according to the IoU between generated curve boxes and labels. Concretely, high-quality, medium-quality and low-quality curve boxes are distinguished by IoU ranges of IoU$\le$0.5, 0.5$\le$IoU$\le$0.7, and IoU$\ge$0.7.}
	\label{V8}
\end{figure*}

\section{Experiments}
\label{Experiments}
In this section, we first introduce the representative text detection datasets and the corresponding evaluation metrics. Then, implementation details about experiments and model settings are illustrated. Next, we explore the proposed edge approximation-based text representation method's fitting ability. Furthermore, the effectiveness of the designed BEP module and PI-loss is demonstrated via ablation studies. Meanwhile, the model performance with different parameter settings of the curve fitting function is illustrated to explore suitable parameters. Ultimately, we show the superior performance of the constructed EdgeText by comparing it with existing state-of-the-art methods on multiple public datasets.

\subsection{Datasets and Evaluation Metrics}
\textbf{Datasets.} Scene texts can be divided into regular and irregular categories according to their shapes. In this paper, we introduce the datasets with multi-oriented rectangle shapes (MSRA-TD500~\cite{yao2012detecting} and ICDAR2015~\cite{karatzas2015icdar}) and curve shapes (Total-Text~\cite{ch2017total} and CTW1500~\cite{yuliang2017detecting}) to evaluate the comprehensive detection performance of our EdgeText on arbitrary-shaped text instances.

\textbf{Evaluation Metrics.} To compare the proposed EdgeText with existing methods for showing superior performance, we follow current popular models to evaluate the performance through the three metrics (Precision, Recall, and H-mean) in object detection. These metrics can be computed as follows:
\begin{eqnarray} 
	\label{e13}
	\begin{gathered}
		\mathrm{Precision}=(\frac{\mathrm{TP}}{\mathrm{TP}+\mathrm{FP}})\times 100\%,\\
		\mathrm{Recall}=(\frac{\mathrm{TP}}{\mathrm{TP}+\mathrm{FN}})\times 100\%,\\
		\mathrm{H-mean}=\frac{2\times \mathrm{Precision}\times \mathrm{Recall}}{\mathrm{Precision}+\mathrm{Recall}},
	\end{gathered}
\end{eqnarray} 
where $\mathrm{TP}$, $\mathrm{FP}$, and $\mathrm{FN}$ are the number of the predicted True Positive, False Positive, and False Negative samples. 

\subsection{Implementation Details}
As introduced in Fig.~\ref{V5} and Section~\ref{Overall Framework}, the EdgeText framework extracts multi-scale features through the feature extractor that comprises a backbone and FPN. In the following experiments, we chose ResNet-50~\cite{he2016deep} that pre-trained on ImageNet~\cite{deng2009imagenet} as the backbone, and the FPN is the version implemented by PyTorch officially.

In the training stage, three kinds of the following data augmentation strategies are adopted to vary the training samples for improving EdgeText's generalization ability: 1) random cropping; 2) random rotating; 3) random scaling; 4) random color distortion. The model is optimized with 1200 epochs under the setting of 16 batch size and 1e-6 initialized learning rate. In the inference stage, the red color arrows in Fig.~\ref{V6} are abandoned, which are responsible for enhancing the model's ability to recognize text edges and truncation point locations in the training stage only. During the training process, we used four 1080ti GPUs to train the model in parallel. For the testing process, a single 1080ti GPU is used to evaluate the model's performance.

\subsection{Fitting Ability of Edge Approximation Representation}
\label{Fitting Ability of Edge Approximation Representation}
An edge approximation-based text representation is proposed to fit arbitrary-shaped scene texts compactly via a simple and intuitive process, which helps avoid the limitations exist in current popular methods. As introduced in Section~\ref{Edge Approximation Representation} and Equation~\ref{e2}, polynomial is adopted as the curve fitting function for approximating text edges. Considering the varied shapes of scene texts, we explore the fitting ability of edge approximation representation on them and find out a suitable parameter settings to obtain a superior comprehensive performance for text instances with different shapes.

Concretely, we analyze the overlap ratio between reconstructed curve boxes based on edge approximation representation and ground-truth contours via the metric of Intersection of Union (IoU) under different settings of fitting function parameters. As shown in Fig.~\ref{V7}, the relationships between the overlap ratio and parameters on multiple public datasets are visualized, where the tick label on the Parameters-axis illustrates the specific parameter setting (e.g., 4(3) means the highest degree $m$ of polynomial is set to 4. Only the parameters of $\{\Theta_i,|i=2,3,4\}$ are adopted to fitting the text edge without the parameters of $\{\Theta_i,c|i=1\}$). For rectangle and quadrilateral boxes labeled datasets (MSRA-TD500~\cite{yao2012detecting} and ICDAR2015~\cite{karatzas2015icdar}), it is found that the proposed edge 
\begin{table}[]
	\centering
	\renewcommand{\arraystretch}{1.2}
	\setlength{\tabcolsep}{1.6mm}
	\caption{Experimental results of EdgeText with different curve fitting function parameter settings on MSRA-TD500 and Total-Text datasets.}
	\begin{tabular}{c|ccc|ccc}
		\Xhline{1pt}
		\multirow{2}{*}{parameter} & \multicolumn{3}{c|}{MSRA-TD500}                                       & \multicolumn{3}{c}{Total-Text}                                       \\ \cline{2-7} 
		& \multicolumn{1}{c}{Precision} & \multicolumn{1}{c}{Recall} & H-mean & \multicolumn{1}{c}{Precision} & \multicolumn{1}{c}{Recall} & H-mean \\ \Xhline{1pt}
		2(1)                       & \multicolumn{1}{c}{91.0}      & \multicolumn{1}{c}{83.7}   & 87.2   & \multicolumn{1}{c}{89.4}      & \multicolumn{1}{c}{85.9}   & 87.6   \\ 
		2(2)                       & \multicolumn{1}{c}{93.3}      & \multicolumn{1}{c}{87.1}   & 90.1   & \multicolumn{1}{c}{88.8}      & \multicolumn{1}{c}{85.1}   & 86.9   \\ 
		2(3)                       & \multicolumn{1}{c}{76.6}      & \multicolumn{1}{c}{81.2}   & 78.8   & \multicolumn{1}{c}{89.1}      & \multicolumn{1}{c}{84.4}   & 86.7   \\ \Xhline{1pt}
	\end{tabular}
	\label{t1}
\end{table}
\begin{figure}
	\centering
	\includegraphics[width=.48\textwidth]{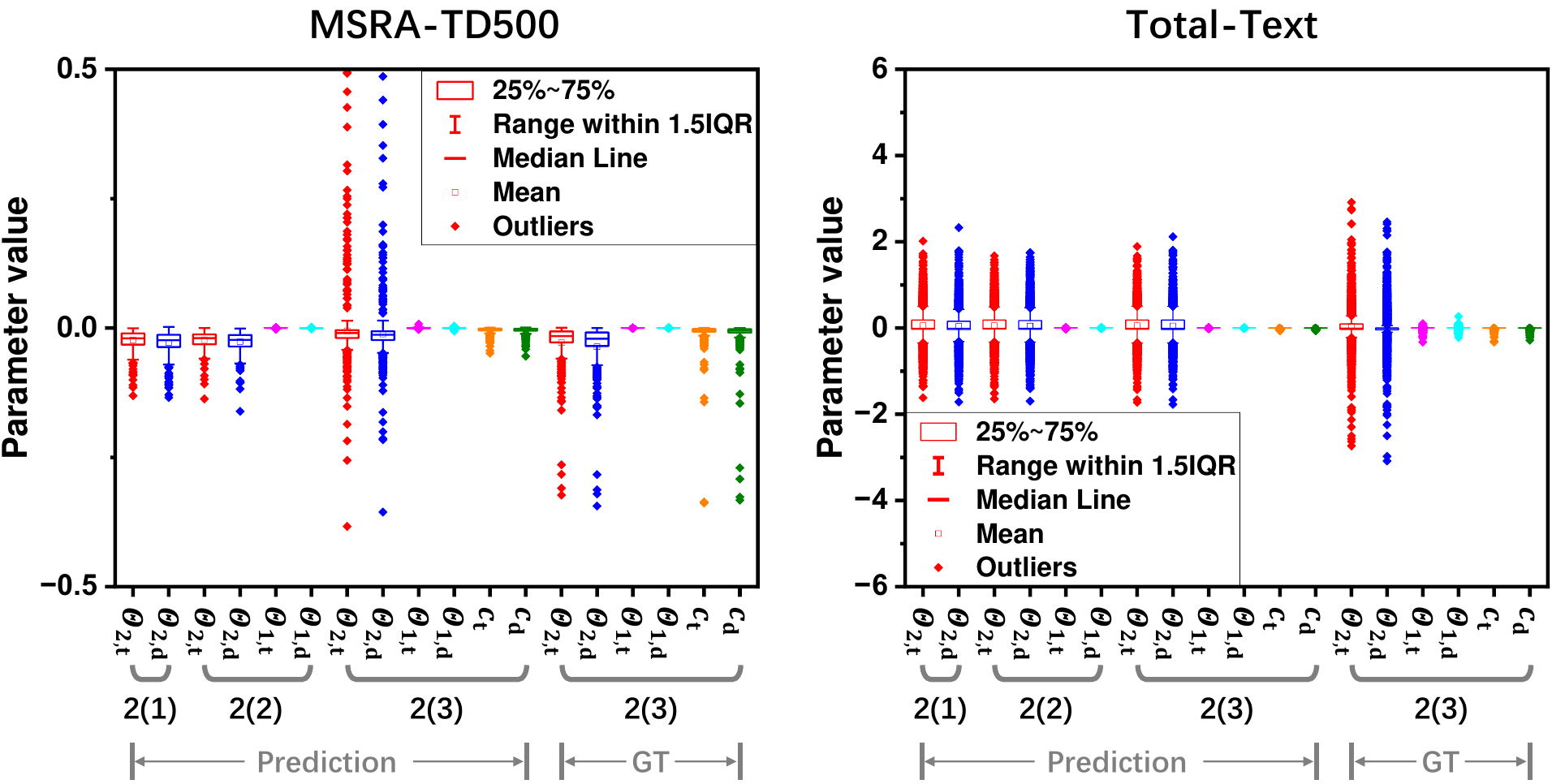}
	\caption{Visualization of the predicted curve fitting function parameters with different settings and the corresponding ground-truth on MSRA-TD500 and Total-Text datasets.}
	\label{V9}
\end{figure}approximation representation enjoys stronger fitting ability with the decrease of the polynomial highest degree (as shown in the Fig.~\ref{V8} first row) and the fitting ability starts to fluctuate around the highest degree is set to 2, which mainly because the overfitting of the high degree polynomial toward edge with simple shapes (such as straight line). The same trend occurs when tuning the number of parameters smaller with a fixed degree (e.g., the ascending curve in the parameters range of 4(5) to 4(1) for MSRA-TD500 and ICDAR2015 datasets). Different from the above two datasets, as visualized in the Fig.~\ref{V8} second row), the text in Total-Text~\cite{ch2017total} and CTW1500~\cite{yuliang2017detecting} is composed of two curved edges that always can be approximated compactly by the polynomial with a higher degree ($m\ge 2$, $m$ denotes the highest degree of polynomial). It would lead to low-quality fitting results when $m\le 1$ because of the intrinsic limitation of straight lines to curves.  Besides, there is an opposite trend compared with MSRA-TD500 and ICDAR2015 datasets in that the overlap ratio decreases when tuning the number of parameters smaller with a fixed degree. It is mainly because some text edge approximating curves are not Y-symmetric and the fitting function with fewer parameters fits these curves worse. 

Overall, based on the above observation from the analysis in Fig.~\ref{V7}, we choose to set the polynomial highest degree $m$ to 2 to ensure a strong comprehensive fitting ability of edge approximation representation toward arbitrary shapes of multiple public datasets.
\begin{table}[]
	\centering
	\renewcommand{\arraystretch}{1.2}
	\setlength{\tabcolsep}{3.5mm}
	\caption{Experimental results of EdgeText with different module and loss function settings on the Total-Text dataset.}
	\begin{tabular}{c|c|ccc}
		\Xhline{1pt}
		\multirow{2}{*}{BEP module} & \multirow{2}{*}{PI-loss} & \multicolumn{3}{c}{Total-Text}                                       \\ \cline{3-5} 
		&                          & \multicolumn{1}{c}{Precision} & \multicolumn{1}{c}{Recall} & H-mean \\ \Xhline{1pt}
		&                          & \multicolumn{1}{c}{86.7}          & \multicolumn{1}{c}{85.4}       &    86.0    \\ 
		$\checkmark$&                          & \multicolumn{1}{c}{88.2}          & \multicolumn{1}{c}{84.9}       &   86.5     \\ 
		$\checkmark$&      $\checkmark$                    & \multicolumn{1}{c}{89.4}          & \multicolumn{1}{c}{85.9}       &    87.6    \\ \Xhline{1pt}
	\end{tabular}
	\label{t2}
\end{table}
\begin{table}[]
	\centering
	\renewcommand{\arraystretch}{1.2}
	\setlength{\tabcolsep}{2.6mm}
	\caption{Experimental results of EdgeText with different truncation point sampling strategies on the Total-Text dataset.}
	\begin{tabular}{c|c|ccc}
		\Xhline{1pt}
		\multirow{2}{*}{From center} & \multirow{2}{*}{From endpoint} & \multicolumn{3}{c}{Total-Text}                                       \\ \cline{3-5} 
		&                                & \multicolumn{1}{c}{Precision} & \multicolumn{1}{c}{Recall} & H-mean \\ \Xhline{1pt}
		$\checkmark$&                                & \multicolumn{1}{c}{88.7}          & \multicolumn{1}{c}{85.8}       &   87.2     \\ 
		&  $\checkmark$                              & \multicolumn{1}{c}{89.4}          & \multicolumn{1}{c}{85.9}       &    87.6    \\ \Xhline{1pt}
	\end{tabular}
	\label{t3}
\end{table}
\begin{table*}[]
	\centering
	\renewcommand{\arraystretch}{1.2}
	\setlength{\tabcolsep}{1.6mm}
	\caption{Performance comparison on multiple public datasets, including MSRA-TD500, ICDAR2015, Total-Text, and CTW1500.}
	\begin{tabular}{c|ccc|ccc|ccc|ccc}
		\Xhline{1pt}
		\multirow{2}{*}{Methods} & \multicolumn{3}{c|}{MSRA-TD500}                                       & \multicolumn{3}{c|}{ICDAR2015}                                        & \multicolumn{3}{c|}{Total-Text}                                       & \multicolumn{3}{c}{CTW1500}                                          \\ \cline{2-13} 
		& \multicolumn{1}{c}{Precision} & \multicolumn{1}{c}{Recall} & H-mean & \multicolumn{1}{c}{Precision} & \multicolumn{1}{c|}{Recall} & H-mean & \multicolumn{1}{c}{Precision} & \multicolumn{1}{c}{Recall} & H-mean & \multicolumn{1}{c|}{Precision} & \multicolumn{1}{c}{Recall} & H-mean \\ \Xhline{1pt}
		EAST'17~~\cite{zhou2017east} & \multicolumn{1}{c}{87.3}          & \multicolumn{1}{c}{67.4}       &    76.1    & \multicolumn{1}{c}{83.3}          & \multicolumn{1}{c}{78.3}       &    80.7    & \multicolumn{1}{c}{49.0}          & \multicolumn{1}{c}{43.1}       &    45.9    & \multicolumn{1}{c}{46.7}          & \multicolumn{1}{c}{37.2}       &    41.4    \\ 
		TextSnake'18~~\cite{long2018textsnake} & \multicolumn{1}{c}{83.2}          & \multicolumn{1}{c}{73.9}       &    78.3    & \multicolumn{1}{c}{84.9}          & \multicolumn{1}{c}{80.4}       &    82.6    & \multicolumn{1}{c}{61.5}          & \multicolumn{1}{c}{67.9}       &    64.6    & \multicolumn{1}{c}{65.4}          & \multicolumn{1}{c}{63.4}       &    64.4    \\ 
		CARFT'19~~\cite{baek2019character} & \multicolumn{1}{c}{88.2}          & \multicolumn{1}{c}{78.2}       &    82.9    & \multicolumn{1}{c}{89.8}          & \multicolumn{1}{c}{84.3}       &    86.9    & \multicolumn{1}{c}{87.6}          & \multicolumn{1}{c}{79.9}       &    83.6    & \multicolumn{1}{c}{86.0}          & \multicolumn{1}{c}{81.1}       &    83.5    \\ 
		TextField'~19~\cite{xu2019textfield} & \multicolumn{1}{c}{87.4}          & \multicolumn{1}{c}{75.9}       &    81.3    & \multicolumn{1}{c}{84.3}          & \multicolumn{1}{c}{83.9}       &    84.1    & \multicolumn{1}{c}{81.2}          & \multicolumn{1}{c}{79.9}       &    80.6    & \multicolumn{1}{c}{83.0}          & \multicolumn{1}{c}{79.8}       &    81.4    \\ 
		ContourNet'~20~\cite{wang2020contournet} & \multicolumn{1}{c}{}          & \multicolumn{1}{c}{}       &        & \multicolumn{1}{c}{87.6}          & \multicolumn{1}{c}{86.1}       &    86.9    & \multicolumn{1}{c}{86.9}          & \multicolumn{1}{c}{83.9}       &    85.4    & \multicolumn{1}{c}{83.7}          & \multicolumn{1}{c}{84.1}       &    83.9   \\ 
		DBNet'20~~\cite{liao2020real} & \multicolumn{1}{c}{91.5}          & \multicolumn{1}{c}{79.2}       &    84.9    & \multicolumn{1}{c}{88.2}          & \multicolumn{1}{c}{82.7}       &    85.4    & \multicolumn{1}{c}{87.1}          & \multicolumn{1}{c}{82.5}       &    84.7    & \multicolumn{1}{c}{86.9}          & \multicolumn{1}{c}{80.2}       &    83.4    \\ 
		CAFA'20~~\cite{dai2019deep} & \multicolumn{1}{c}{}          & \multicolumn{1}{c}{}       &        & \multicolumn{1}{c}{86.2}          & \multicolumn{1}{c}{82.7}       &    84.4    & \multicolumn{1}{c}{84.6}          & \multicolumn{1}{c}{78.6}       &    81.5    & \multicolumn{1}{c}{85.7}          & \multicolumn{1}{c}{85.1}       &    85.4    \\ 
		FCENet'21~~\cite{zhu2021fourier} & \multicolumn{1}{c}{}          & \multicolumn{1}{c}{}       &        & \multicolumn{1}{c}{90.1}          & \multicolumn{1}{c}{82.6}       &    86.2    & \multicolumn{1}{c}{89.3}          & \multicolumn{1}{c}{82.5}       &    85.8    & \multicolumn{1}{c}{87.6}          & \multicolumn{1}{c}{83.4}       &    85.5    \\ 
		PCR'21~~\cite{dai2021progressive} & \multicolumn{1}{c}{90.8}          & \multicolumn{1}{c}{83.5}       &    87.0    & \multicolumn{1}{c}{}          & \multicolumn{1}{c}{}       &        & \multicolumn{1}{c}{88.5}          & \multicolumn{1}{c}{82.0}       &    85.2    & \multicolumn{1}{c}{87.2}          & \multicolumn{1}{c}{82.3}       &    84.7    \\ 
		ABCNet'21~~\cite{liu2021abcnet} & \multicolumn{1}{c}{89.4}          & \multicolumn{1}{c}{81.3}       &    85.2    & \multicolumn{1}{c}{90.4}          & \multicolumn{1}{c}{86.0}       &    88.1    & \multicolumn{1}{c}{90.2}          & \multicolumn{1}{c}{84.1}       &    87.0    & \multicolumn{1}{c}{85.6}          & \multicolumn{1}{c}{83.8}       &    84.7    \\ 
		CM-Net'22~~\cite{DBLP:journals/tip/YangCXYW22} & \multicolumn{1}{c}{89.9}          & \multicolumn{1}{c}{80.6}       &    85.0    & \multicolumn{1}{c}{86.7}          & \multicolumn{1}{c}{81.3}       &    83.9    & \multicolumn{1}{c}{88.5}          & \multicolumn{1}{c}{81.4}       &    84.8    & \multicolumn{1}{c}{86.0}          & \multicolumn{1}{c}{82.2}       &    84.1    \\ 
		NASK'22~~\cite{cao2021all} & \multicolumn{1}{c}{}          & \multicolumn{1}{c}{}       &        & \multicolumn{1}{c}{90.9}          & \multicolumn{1}{c}{89.2}       &    90.0    & \multicolumn{1}{c}{85.6}          & \multicolumn{1}{c}{83.2}       &    84.4    & \multicolumn{1}{c}{83.4}          & \multicolumn{1}{c}{80.1}       &    81.7    \\ 
		RFN'22~~\cite{guan2022industrial} & \multicolumn{1}{c}{88.4}          & \multicolumn{1}{c}{87.8}       &    88.1    & \multicolumn{1}{c}{}          & \multicolumn{1}{c}{}       &        & \multicolumn{1}{c}{}          & \multicolumn{1}{c}{}       &        & \multicolumn{1}{c}{}          & \multicolumn{1}{c}{}       &        \\ 
		ASTD'22~~\cite{zhang2022arbitrary} & \multicolumn{1}{c}{88.6}          & \multicolumn{1}{c}{82.7}       &    85.6    & \multicolumn{1}{c}{88.2}          & \multicolumn{1}{c}{83.3}       &    85.7    & \multicolumn{1}{c}{89.8}          & \multicolumn{1}{c}{84.8}       &    87.2    & \multicolumn{1}{c}{87.6}          & \multicolumn{1}{c}{80.8}       &    84.1    \\ 
		DeepSolo'23~~\cite{ye2023deepsolo} & \multicolumn{1}{c}{}          & \multicolumn{1}{c}{}       &        & \multicolumn{1}{c}{92.8}          & \multicolumn{1}{c}{87.4}       &    \textbf{90.0}    & \multicolumn{1}{c}{93.9}          & \multicolumn{1}{c}{82.1}       &    87.3    & \multicolumn{1}{c}{}          & \multicolumn{1}{c}{}       &        \\ 
		LeafText'23~~\cite{yang2023text} & \multicolumn{1}{c}{92.1}          & \multicolumn{1}{c}{83.8}       &    87.8    & \multicolumn{1}{c}{88.9}          & \multicolumn{1}{c}{82.3}       &    86.1    & \multicolumn{1}{c}{90.8}          & \multicolumn{1}{c}{84.0}       &    87.3    & \multicolumn{1}{c}{87.1}          & \multicolumn{1}{c}{83.9}       &    85.5    \\ CBNet'24~~\cite{zhao2024cbnet} & \multicolumn{1}{c}{91.1}          & \multicolumn{1}{c}{84.8}       &    87.8    & \multicolumn{1}{c}{89.0}          & \multicolumn{1}{c}{95.5}       &    87.2    & \multicolumn{1}{c}{90.1}          & \multicolumn{1}{c}{82.5}       &   86.1    & \multicolumn{1}{c}{89.0}          & \multicolumn{1}{c}{81.9}       &    85.3    \\  
		VTD'24~~\cite{zhang2023video} & \multicolumn{1}{c}{89.2}          & \multicolumn{1}{c}{81.5}       &    85.2    & \multicolumn{1}{c}{88.5}          & \multicolumn{1}{c}{85.8}       &    87.1    & \multicolumn{1}{c}{}          & \multicolumn{1}{c}{}       &        & \multicolumn{1}{c}{}          & \multicolumn{1}{c}{}       &        \\ \hline
		\textbf{Ours}& \multicolumn{1}{c}{93.3}          & \multicolumn{1}{c}{87.1}       &    \textbf{90.1}    & \multicolumn{1}{c}{89.8}          & \multicolumn{1}{c}{83.6}       &     86.6   & \multicolumn{1}{c}{89.4}          & \multicolumn{1}{c}{85.9}       &     \textbf{87.6}   &  \multicolumn{1}{c}{86.9}          & \multicolumn{1}{c}{84.3}       &    \textbf{85.6}   \\ \Xhline{1pt}
	\end{tabular}
	\label{t4}
\end{table*}

\begin{figure*}[h]
	\centering
	\includegraphics[width=.95\textwidth]{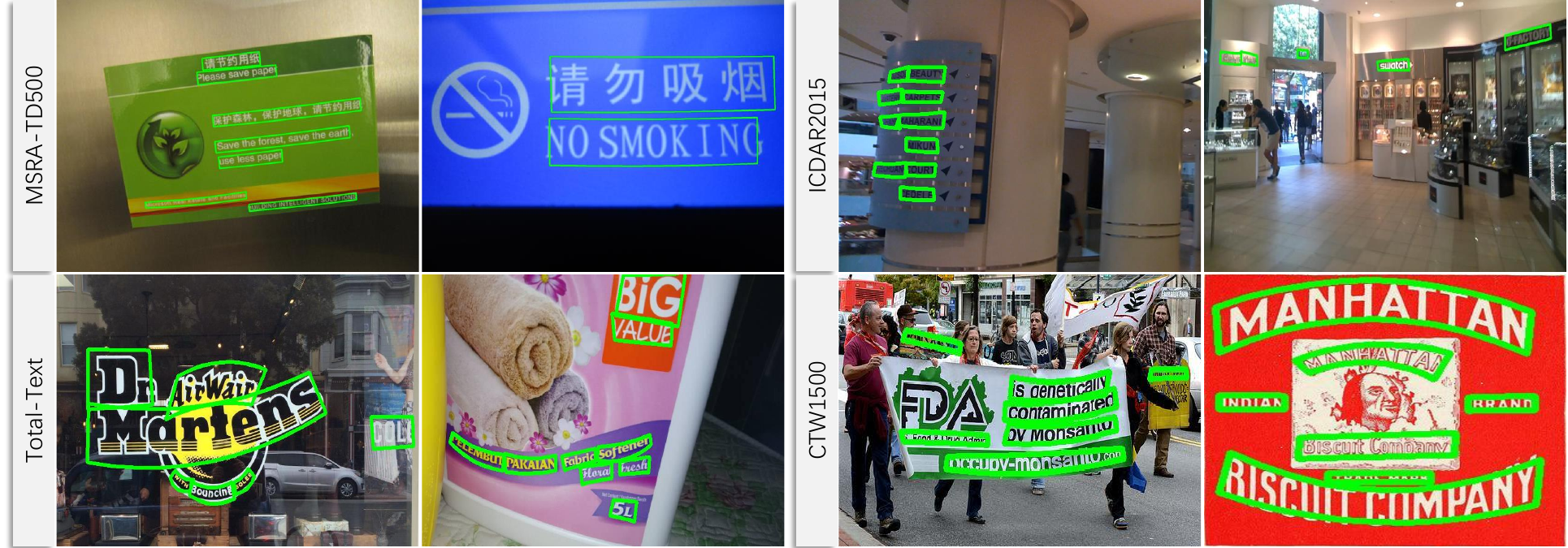}
	\caption{Visualization of some quality detection results on regular (MSRA-TD500 and ICDAR2015) and irregular-shaped (Total-Text and CTW1500) text detection datasets. It can be observed that the proposed method can fit arbitrary-shaped scene texts simultaneously and effectively.}
	\label{V10}
\end{figure*}

\subsection{Ablation Study}
\label{Ablation Study}
In the above Section~\ref{Fitting Ability of Edge Approximation Representation}, the strong fitting ability of the proposed edge approximation representation toward regular and irregular-shaped texts has been shown and suitable parameter settings ($m=2$) of the curve fitting function have been determined. We will further explore the performance influences brought by different numbers of parameters under the setting of $m=2$ and verify the effectiveness of the introduced BEP module and PI-loss.

\begin{figure*}
	\centering
	\includegraphics[width=.95\textwidth]{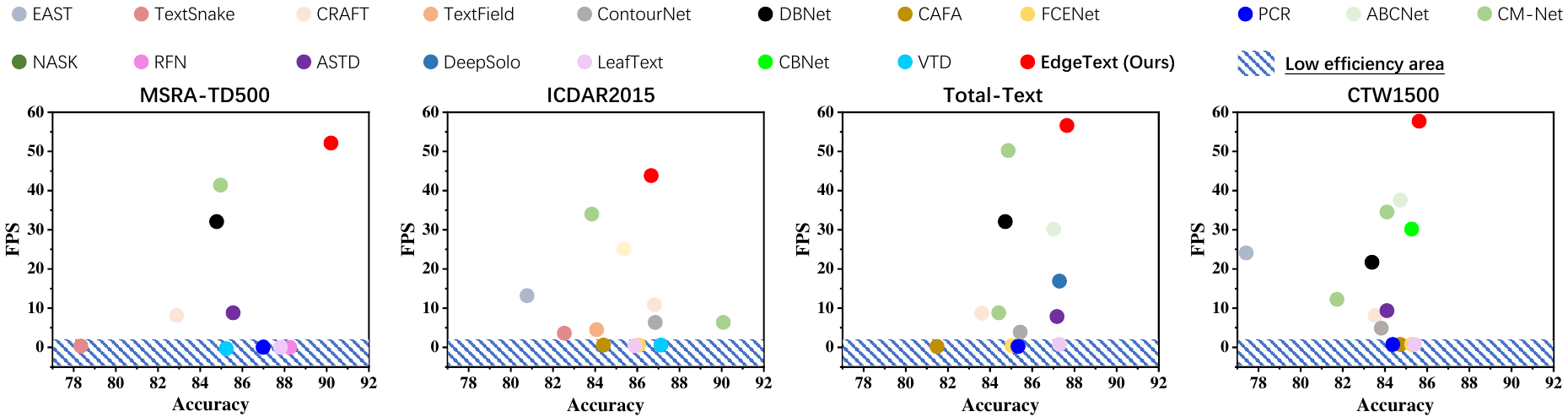}
	\caption{Visualization of the comparison between the efficiency and accuracy balance with existing advanced methods.}
	\label{V11}
\end{figure*}
\begin{figure}
\centering
\includegraphics[width=.49\textwidth]{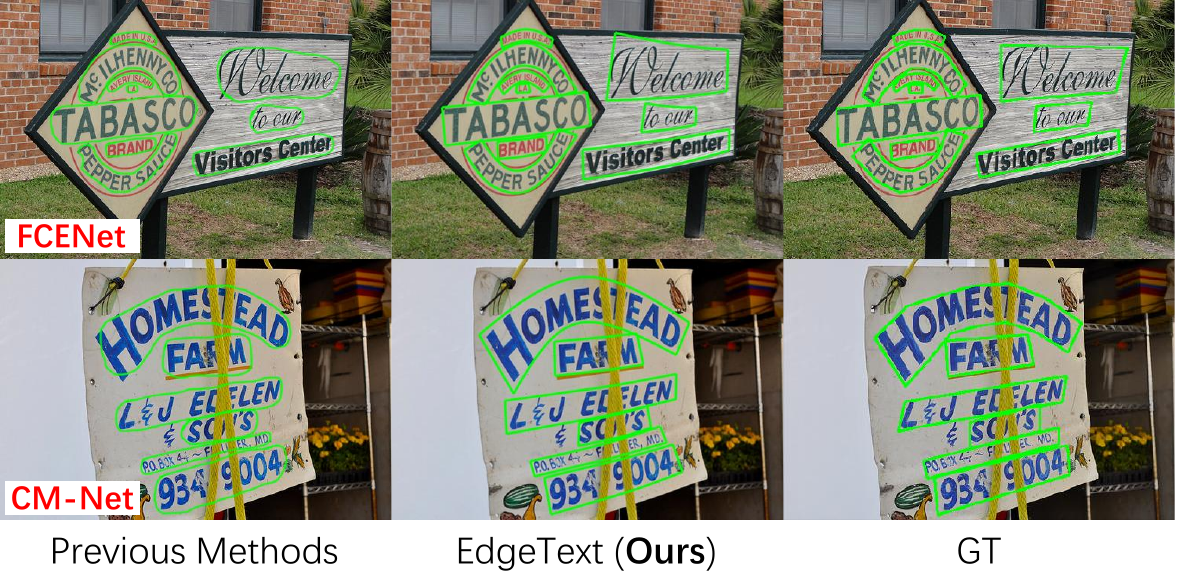}
\caption{Qualitative comparisons with FCENet~\cite{zhu2021fourier} and CM-Net~\cite{DBLP:journals/tip/YangCXYW22} on some challenging samples. FCENet and CM-Net are the representative irregular-shaped text detection methods that adopt the piecewise fitting and shrink-mask expand strategies, respectively.}
\label{V12}
\end{figure}
\textbf{Influence of the number of parameters.} As shown in Table~\ref{t1}, EdgeText achieves comparable performance in H-mean (87.2\% and 90.1\%) when building the curve fitting function without constant term on MSRA-TD500. Meanwhile, the model performance degrades significantly when reconstructing with constant terms in the inference stage. It is observed that the predicted $\Theta_{2,t}$ and $\Theta_{2,d}$ of the function deviates far from the corresponding ground truth when the constant term participates in the training process (as shown in the parameter analysis in left sub-figure of Fig.~\ref{V9}). Since $\Theta_{2,t}$ and $\Theta_{2,d}$ are more important for rebuilding contours than other lower-degree parameters and constant terms, the parameter deviation results in performance degradation. On the Total-Text dataset, the performance shows a slight decrease trend with the number of parameters tuned larger. Especially, there isn't a significant performance decrease trend when the constant term is considered in model optimization. It is mainly because the corresponding label distributes around 0, which brings limited influence to the model to learn coefficients of a higher polynomial degree.

\textbf{Effectiveness of BEP module and PI-loss.} Considering the deep dependency of EdgeText on text edges, the BEP module and PI-loss are introduced to encourage our model to recognize edge features more accurately. As shown in Table~\ref{t2}, compared with the baseline, the BEP module and PI-loss bring 0.5\% and 1.1\% improvements in H-mean. Specifically, the above two components help our model recognize the text edge features effectively in the optimization process, which enhances the precision of the rebuilt text contours in the inference stage (1.5\% and 1.2\% improvements in Precision metric). The above experimental results demonstrate the effectiveness of the BEP module and PI-loss in improving EdgeText's ability to recognize text edges more accurately.

\textbf{Influence of different truncation point sampling strategies.} As described in Section~\ref{Overall Framework}, considering the large offsets between center points and truncation points, the lateral vein prediction strategy proposed in LeafText~\cite{yang2023text} is adopted to determine the truncation points. It allows the proposed EdgeText to determine the truncation point locations by predicting shorter offsets between truncation points and center line endpoints of concentric masks instead of the large offsets between truncation points and centers, which helps predict truncation points more accurately. As shown in Table~\ref{t3}, the strategy proposed in LeafText brings 0.4\% performance improvements in H-mean compared with determining the truncation points by combining the text centers and the offsets between truncation points and centers.

\subsection{Comparison Experiments}
According to the experimental analysis in Section~\ref{Fitting Ability of Edge Approximation Representation} and Section~\ref{Ablation Study}, the effectiveness of the proposed BEP module and PI-loss are verified and suitable model settings are found for obtaining a superior comprehensive performance on varied texts. Based on the above results, we compare EdgeText with existing approaches for showing the superiority of our model.

As shown in Table~\ref{t4} and Fig.~\ref{V10}, the detection results of our model on various public datasets are illustrated. For large-scale text instances in MSRA-TD500 dataset, EdgeText achieves 90.1\% H-mean, which outperforms the existing state-of-the-art (SOTA) method (LeafText~\cite{yang2023text}) 2.3\%. As described in Section~\ref{Methodology}, we adopt the lateral vein prediction strategy proposed in LeafText to encourage determining the truncation points more accurately. Furthermore, compared with the piecewise fitting process of LeafText, our method represents the text contour as a whole, which avoids the cumulative error that may exist in a gradual rebuilding process and ensures the compactness of reconstructed contours. Benefiting from the above advantages, EdgeText outperforms existing methods a lot. For ICDAR2015, a small-scale regular text dataset compared with MSRA-TD500, our model behind of DeepSolo~\cite{ye2023deepsolo} slightly. It is mainly because the concentric mask-based text location strategy proposed in CM-Net~\cite{DBLP:journals/tip/YangCXYW22} makes it hard to determine small instances because of the lesser pixel-level positive samples. Besides of the above regular-shaped text datasets, EdgeText enjoys superior performance on irregular datasets (Total-Text and CTW1500). Concretely, it surpasses LeafText~\cite{yang2023text} 0.3\% in H-mean on Total-Text dataset and ASTD~\cite{zhang2022arbitrary} 1.5\% in H-mean on CTW1500. The above superiority of our model brought by the edge approximation representation method that fits texts compactly and continuously in an intuitive way, which helps EdgeText detect texts more accurately. Furthermore, in Fig.~\ref{V12}, we also compare the compactness of the detection results from EdgeText and existing representative methods that adopt piecewise fitting and shrink-mask expand strategies, respectively. Compared with previous methods, the compact contours generated by our method effectively avoid bringing unnecessary visual information into the subsequent text recognition process. This helps to reduce background interference and significantly improves overall recognition accuracy.

\subsection{Efficiency Analysis}
As described in Section~\ref{Edge Approximation Representation}, our proposed edge approximation representation method addresses the arbitrary-shaped text detection problem as a curve box regression process. This approach enables the parallel rebuilding of text instances with varying scales and shapes, thereby simplifying post-processing procedures compared to existing advanced methods and reducing inference time. As illustrated in Fig.~\ref{V11}, our method outperforms previous approaches in terms of speed on multiple representative public datasets while maintaining competitive accuracy. The efficiency of the detection process is largely attributed to the streamlined post-processing steps. Additionally, the designed Bilateral Enhanced Perception (BEP) module facilitates the extraction of strong, representative features with a lightweight network, ensuring high detection accuracy at a lower resource cost. These results underscore the superior efficiency of our model.

\section{Conclusion}
\label{Conclusion}
In this paper, we design an edge approximation representation based on the observation of scene text shapes and construct a novel text detection framework, namely EdgeText. The proposed representation helps fit arbitrary-shaped texts compactly and continuously, which avoids the intrinsic limitations that exist in current popular methods. Meanwhile, the edge approximation representation allows our model to rebuild multiple text contours in parallel, which helps EdgeText enjoy a more intuitive contour rebuilding process with fewer procedures compared with existing box-to-polygon strategy-based or piecewise fitting methods. Furthermore, considering the deep dependencies of EdgeText on the text edges and truncation points, the bilateral enhanced perception (BEP) module is designed to improve the model's ability to recognize edge features. In the end, we introduce a proportional integral loss (PI-loss) for accelerating the parameters prediction of the edge approximating curve functions. Ablation studies demonstrate the effectiveness of the introduced BEP module and PI-loss. Comparison experiments on the multiple datasets show the superior performance of our EdgeText.

\bibliographystyle{IEEEtran}
\bibliography{egbib}

\newpage

\vfill

\end{document}